%% file: samplepaper.tex
\documentclass[runningheads]{llncs}

\usepackage{amsmath,amssymb,amsfonts} 
\usepackage{graphicx}
\usepackage{booktabs}
\usepackage{setspace}
\usepackage{hyperref,cleveref}
\usepackage{listings}
\usepackage{esvect}
\usepackage{xspace}
\usepackage{url}

\usepackage{mathrsfs}
\usepackage{enumerate}
\usepackage{nicefrac} 
\usepackage{microtype}
\usepackage{makecell}
\usepackage{lscape}
\usepackage{xspace}

\usepackage{esvect}
\usepackage{mathrsfs}
\usepackage{enumerate}
\usepackage{nicefrac}
\usepackage{lscape}

\usepackage{algorithm}
\usepackage[english]{babel}
\usepackage[utf8]{inputenc}
\usepackage[noend]{algpseudocode}

\usepackage{todonotes}
\usepackage{paralist}
\usepackage{tkz-graph}
\usepackage{tikz}
\usetikzlibrary{arrows,topaths,calc}
\usetikzlibrary{positioning}
\usetikzlibrary{shadows}% =>https://www.overleaf.com/learn/how-to/SyncTeX_Errors.
\usetikzlibrary{shapes}
\usetikzlibrary{backgrounds}
\usetikzlibrary{decorations.pathreplacing}

\usepackage[nolist]{acronym}
\input{acronyms.tex}
\input{notation.tex}

\begin{document}

\title{\approach -- Deep Reinforcement Learning for Refinement Operators in \ALC}
\author{Caglar Demir \and Axel-Cyrille Ngonga Ngomo}
\institute{Data Science Research Group, Paderborn University}
%\author{XXX XXX \and XXX XXX}
%\authorrunning{X}
\maketitle

%%%%%%%%%%%%%%%% ABSTRACT BEGINS %%%%%%%%%%%%%%%%
\begin{abstract}
Approaches based on refinement operators have been successfully applied to class expression learning on RDF knowledge graphs. These approaches often need to explore a large number of concepts to find adequate hypotheses. This need arguably stems from current approaches relying on myopic heuristic functions to guide their search through an infinite concept space. In turn, deep reinforcement learning provides effective means to
address myopia by estimating how much discounted cumulated future reward states promise. In this work, we leverage deep reinforcement learning to accelerate the learning of concepts in \ALC by proposing \approach---a novel class expression learning approach that uses a convolutional deep Q-learning model to steer its search. By virtue of its architecture, \approach is able to compute the expected discounted cumulated future reward of more than $10^3$ class expressions in a second on standard hardware. We evaluate \approach on four benchmark datasets against state-of-the-art approaches. Our results suggest that \approach converges to goal states at least 2.7$\times$ faster than state-of-the-art models on all benchmark datasets. We provide an open-source implementation of our approach, including training and evaluation scripts as well as pre-trained models.\footnote{\url{https://github.com/dice-group/DRILL}}
\end{abstract}
%%%%%%%%%%%%%%%% Introduction BEGINS %%%%%%%%%%%%%%%%
\section{Introduction}
\label{sec:introduction}
\acp{KG} represent structured collections of facts describing the world in the form of typed relationships between entities~\cite{hogan2020knowledge}. These collections of facts have been used in a wide range of applications including web search~\cite{eder2012knowledge}, question answering~\cite{bordes2014question}, recommender systems~\cite{zhang2016collaborative}, cancer research~\cite{saleem2014big}, machine translation~\cite{moussallem2019}, and even entertainment~\cite{malyshev2018getting}.

% What are we concerned with ? and  Why is it important important?
In this work, we focus on the problem of~\ac{CEL} on RDF~\acp{KG} as defined in~\cite{lehmann2010concept}.~\ac{CEL} refers to the problem of learning~\ac{OWL} class expressions from a given background knowledge and examples~\cite{lehmann2010learning}. Tackling \ac{CEL} is particularly important as learned~\ac{OWL} class expressions can be transformed into natural language through verbalization~\cite{ngonga2019holistic} which makes predictions \textit{ante-hoc explainable}. Ergo, optimizing \ac{CEL} has the potential of easing the use of explainable AI in real-life  applications along with the corresponding societal advantages tied to explainability~\cite{burnett2020explaining,samek2019explainable}. The lack of transparency and explainability in AI systems reduces the trust and the verifiability of the decisions made~\cite{samek2017explainable,holzinger2019causability,doshi2017towards,samek2019towards}.

% What is the definition of CEL? and How do state-of-the-art models tackle CEL?
The formal setting for \ac{CEL} is as follows: Given background knowledge \kg, a set of positive $E^+$, and a set of negative examples $E^-$ as well as a formal logic \lang, 
the goal is to find \textbf{a class expression} $H$ in \lang such that $\kg \models H(p) \wedge \kg \models \neg H(n)$, where $p \in E^+$ and $n \in E^-$. This learning problem is often formulated as a search problem in a quasi-ordered state space$(\StateSpace,\preceq)$~\cite{buhmann2016dl,lehmann2011class,fanizzi2008dl,fanizzi2019boosting}. Within this setting, the CEL problem is tackled by learning a sequence of states that starts from an initial state (e.g., $\top$) and leads to a $H$, e.g., $\texttt{Brother} \sqcup \texttt{Sister}$ in~\Cref{example:1}.
\begin{example}
Given a set of individuals who have siblings as $E^+$ and a set of only-children as $E^-$, a CEL approach could aim to learn $\texttt{Brother} \sqcup \texttt{Sister}$.
\label{example:1}
\end{example}
The search of $H$ is steered by optimizing a heuristic function that aims to estimate how likely a state $s_i \in \StateSpace$ leading to $H$, ergo the heuristic value signals how well a class expression fits a learning problem and can be used to guide the search~\cite{lehmann2010learning,lehmann2011class}. % For many learning problems, finding $H \in \hypotheses$ is not possible or too costly (especially in terms of runtime)~\cite{lehmann2010learning,bin2016towards}. 
% CEL is hence often reduced to finding class expressions which maximize an objective function (e.g., the F-measure)~\cite{lehmann2011class}. 
State-of-the-art approaches rely on heuristic functions that determine a heuristic value of $s_i$ without any consideration for future states (see~\Cref{sec:preliminaries}). This is akin to making decisions in a chess game solely based on the current board state and without any consideration for next possible board configurations. We argue that this is an important drawback of current \ac{CEL} models, which leads to state-of-the-art models often needing to explore a large number of states to find satisfactory class expressions~\cite{bin2016towards}. To remedy the problem of exploring a large number of states, a common practice is to apply (1) numerous handcrafted rules and/or (2) a reasoner to remove redundant states from the search, e.g., replace $\forall r. \top$ with $\top$~\cite{lehmann2010learning}. However this treatments result in increasing runtime requirements. Arguably, using such heuristic functions, handcrafted rules or reasoners hinders tackling \ac{CEL} problem in large setting due to increased memory or runtime requirements.

% What do we do in this work ?
In this work, we regard the problem of finding a hypothesis $H \in \hypotheses$ as a sequential decision-making problem and
formalize it within the setting of \ac{RL}. Therein, state-of-the-art \ac{CEL} approaches are analogous to myopic \ac{RL} agents as their objectives are set to solely maximize immediate rewards (see~\Cref{section:methodology}). In contrast, deep \ac{RL} is designed to address the problem of learning how to incorporate consideration for future states in immediate actions~\cite{sutton2018reinforcement,mnih2015human}. Thus, we propose a convolution deep Q-learning model (dubbed \approach) to efficiently steer the search towards $\hypotheses$, while incorporating consideration for future states in immediate decisions. Hence, the problem of finding a hypothesis is formalized as training \approach to select actions in a fashion that maximizes \emph{cumulative discounted future rewards}.

% Unsupervised training.
To train \approach, we adapt the idea of playing Atari 2600 games and design an unsupervised training procedure in a fashion akin to~\cite{mnih2015human}. Our approach to training ensures that~\approach can be trained on \acp{KG} with no adjustment of the architecture. For a given learning problem, we first construct a quasi-ordered state space as a \ac{RL} environment, where each \ac{RL} state is represented with continuous vector representations (\textit{embeddings}) of individuals belonging to the respective \ac{OWL} class expression. Next, \approach begins to interact with this environment by refining the most general state $\rho(\top)$. Subsequently, \approach selects next possible states $s_i \in \rho(\top)$ via $\epsilon-$greedy fashion (see~\Cref{section:methodology}). Thereupon, \approach receives the reward for transitioning from $s_i$ to the next state $s_j$ that it selects.

% What are our contributions?
Our experiments are carried out on the description logic $\mathcal{ALC}$ and show that \approach is able to steer the search for elements of \hypotheses more efficiently than state-of the art approaches. \approach finds goal states \textbf{at least 2.7 times faster} than state-of-the-art approaches on benchmark datasets. Importantly, \approach is able to estimate heuristic values of more than $10^3$ states in a second on standard hardware. The main contributions of this paper are threefold:
\begin{enumerate}
    \item We model \ac{CEL} using refinement operators within the framework of \ac{RL}.
    \item We present a convolutional model for predicting the cumulative discounted future reward for states within an infinite state space. 
    \item We provide an open-source implementation of our framework to foster research in the direction of combining reinforcement learning with class expression learning.
\end{enumerate}
%%%%%%%%%%%%%%%% Related BEGINS %%%%%%%%%%%%%%%%
\section{Related Work}
\label{sec:relatedwork}
Our work is grounded in two areas of research, i.e., \ac{CEL} and \ac{RL}.  
One of the first works of supervised learning in description logic is presented in~\cite{Cohen1992ComputingLC}. Later, Badea et al.~\cite{BadeaShan:2000} propose to apply a refinement operator in a top-down fashion. YINYANG~\cite{iannone2007algorithm} effectively combines the previous approaches to learn class expressions. OCEL~\cite{lehmann2010learning}, ELTL and CELOE~\cite{lehmann2011class} use a proper and complete refinement operator to build a search tree of \ac{OWL} class expressions. Therein, each node corresponds to an \ac{OWL} class expression and is annotated with a respective heuristic and quality values.  In~\Cref{subsec:cel_in_dl}, we elucidate the working of OCEL, ELTL and CELOE. In turn, DL-FOIL~\cite{fanizzi2008dl} proposes to use unlabeled individuals in its heuristic function to take the open world assumption into account. 
% Commented our as one of reviewer understood this statement as if it is ours.
% Findings presented in~\cite{fanizzi2008dl,fanizzi2018dlfoil,fanizzi2019boosting} suggest that DL-FOIL is more effective than YINYANG and CELOE at approximating target complex class expressions.

Over the past decade, there have been a number of successes in learning policies for sequential decision-making problems. Notable examples include deep Q-learning for Atari game-playing~\cite{mnih2015human} and strategic policies joined with search led to defeating a human expert at the game of Go~\cite{Silver16Go}. Likewise, various~\ac{RL} models have been applied in diverse tasks on \acp{KG}, including question answering, link prediction, fact checking and knowledge graph completion~\cite{zheng2018drn,xian2019reinforcement,das2017walk,lin2018multi,xiong2017deeppath}. % Such~\ac{RL} models often presuppose that the number of actions are known in advance. In our work, this presupposition is untenable as the number of possible actions depends a given state.
%No previous works has combined the two family of approaches. In this paper, we use \ac{RL} to improve \ 

\section{Preliminaries}
\label{sec:preliminaries}
\subsection{Class Expression Learning in Description Logics}
\label{subsec:cel_in_dl}
We define the problem of class expression learning in a fashion akin to~\cite{lehmann2011class,lehmann2010learning,lehmann2010concept,lehmann2014concept}. Given a $\kg=(\TBox,\ABox)$, a set of positive $E^+$ and negative examples $E^-$ as well as a formal logic \lang (e.g., \ALC),  the goal is to find a class expression $H \in \hypotheses$ defined as
\begin{equation}
  \forall \, H \in \hypotheses: \{ \big(\kg \models H(p)) \wedge 
  (\kg \models \neg H(n) \big)\} \, \text{s.t.} \, \forall p \in E^+, \forall n \in E^-.
  \label{eq:cel}
\end{equation}

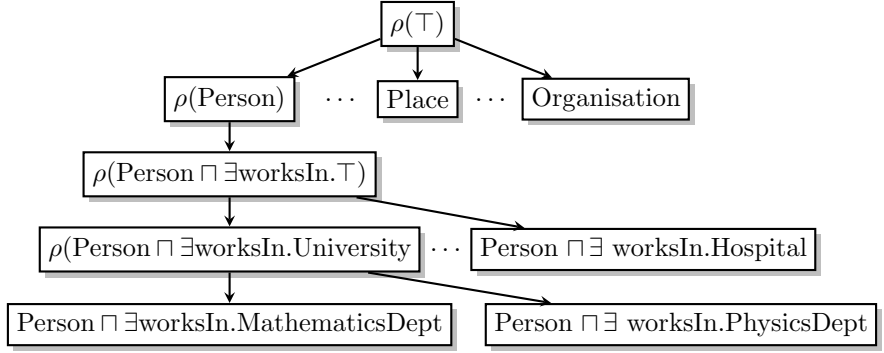
\begin{figure*}[htb]
    \centering
    \input{Images/SeachSpace.tikz}
    \caption{Illustration of traversing $\StateSpace$ and building a search tree in $\mathcal{ALC}$}
    \label{fig:traverse_search_space}
\end{figure*}

The problem of finding hypotheses is transformed into a search problem within a quasi-ordered state space $(\StateSpace,\preceq)$~\cite{lehmann2011class}, where each state corresponds to a class expression. A quasi-ordering imposes a reflexive and transitive relation between states. The quality of a given class expression is often determined by means of accuracy or F1-measure~\cite{lehmann2011class}. Traversing in \StateSpace is commonly conducted via top-down refinement operators $\rho:\StateSpace \rightarrow 2^\StateSpace$~\cite{lehmann2007refinement}:
\begin{equation}
\label{eq:refinement}
     \forall s_i \in \StateSpace : \rho(s_i) \subseteq \{ s_j \in \mathcal{S} \;|\; s_j \preceq s_i \},
\end{equation}
where $\rho(s_i)$ yields a set of \emph{specialisations} of $s_i$. 
% Bottom-up refinement operators are defined analogously~\cite{lehmann2011class}. 
The search tree is initialized via adding the most general state ($\top$) as a root node. The initialized search tree is iteratively built by selecting a node containing the most promising state and adding its qualifying refinements as its children into a search tree. ~\Cref{fig:traverse_search_space} illustrates an excerpt of traversing $\StateSpace$ and building a search tree for a learning problem on a fictitious university ontology.
% An excerpt of how a top-down refinement operator traverses the space of \ALC class expressions of an exemplary knowledge base is shown in~\Cref{fig:traverse_search_space}. 
%
% The search of $\hypotheses$ begins with the refinement of the most general state $\rho(\top)$ and is steered by a heuristic function that quantifies the promise of a state leading to $H \in \hypotheses$. Assume that $\texttt{Person} \, \sqcap \, \exists \texttt{ worksIn.MathematicsDept} \in \hypotheses$, then finding a goal state can be considered as taking the following sequence of actions, i.e. refining states; $\big[ \rho(\top), \rho(\texttt{Person} \sqcap \exists \texttt{worksIn.}\top), \rho(\texttt{Person} \sqcap \exists \texttt{worksIn.University})) \big]$.
%
\subsection{Heuristic functions} 
A major challenging part of this search problem is to find  a suitable heuristic function that efficiently steers the search towards elements of $\hypotheses$~\cite{lehmann2011class}. A heuristic function quantifies how adequately a class expression fits a learning problem and guides the search in a learning process~\cite{lehmann2011class}. 

To introduce heuristic functions of state-of-the-art approaches, we first define the retrieval function. Let $\Classes,\Individuals$ be all class expressions in \lang and all individuals in $\ABox$, respectively. We follow Definition 2.17 in~\cite{lehmann2010learning} and define the retrieval function as $\Retrievalfunc : \; \Classes \mapsto \Individuals$. $\Retrievalfunc$ maps class expressions to the set of its instances, e.g. $\Retrievalfunc(\texttt{Person})$ will contain all instances of the class \texttt{Person}. The heuristic of OCEL is defined as 
\begin{equation}
    \heuristic_\text{OCEL}(c_i,c_j) = \text{A}(c_j) + \lambda \cdot \big[ \text{A}(c_j)-\text{A}(c_i) \big] - \beta \cdot z,
\label{eq:ocel}    
\end{equation}
where $\beta > \lambda \geq 0$ and $z \in \NaturalNumber$ stands for the horizontal expansion value~\cite{lehmann2010learning} and $A$ is defined as 
\begin{equation}
\text{A}(c) = 1- \frac{|E^+ \setminus \Retrievalfunc(c)| + | \Retrievalfunc(c) \cap E^- | }{|E^+| + |E^-|}.
    \label{eq:acc}
\end{equation}
$A(c_i)$ quantifies the quality of a class expression $c_i$ w.r.t.$E^+$ and $E^-$, whereas 
$\heuristic_\text{OCEL}(c_i,c_j)$ determines the heuristic value of $c_j$ given $c_i$ w.r.t. $E^+$ and $E^-$. ELTL replaces the stepwise horizontal expansion technique used in OCEL with a penalty for the length of a concept. CELOE extends OCEL and ELTL by redefining the accuracy of a class expression and replacing $z$ with a length bias (see the definition 5.4 in \cite{lehmann2010learning}) to incorporate the number of all individuals in the computation of heuristic values. The heuristic function of CELOE is defined as 
\begin{equation}
\heuristic_\text{CELOE}(c_i,c_j) = \text{A}(c_j) + \lambda \cdot \big[ \text{A}(c_j)-\text{A}(c_i) \big] - \beta \cdot |c_j|,\mbox{ where}
\label{eq:celoe}
\end{equation}

\begin{equation}
\text{A} (c,t) = 1 - 2 \cdot \frac{t \cdot |E^+ \setminus \Retrievalfunc(c)|
+ | \Retrievalfunc(c) \setminus E^+ |}{ (t+1) \cdot |\Individuals|}.
\label{eq:celoe_acc}    
\end{equation}
$t>1$ and $|c_j|$ is the length of $c_j$.
% However, the rationale remains the same; at each time step, 
% (1) select the most promising class expression on the search tree, that is determined by
% $\heuristic_\text{CELOE}$ or $\heuristic_\text{OCEL}$, (2) add qualifying refinements of the most promising class expression into the search tree.
Note that for the sake of brevity, we do not introduce a new symbol for a node containing a class expression with a score and horizontal expansion in \Cref{eq:ocel} and \Cref{eq:celoe}.
\subsection{Reinforcement Learning}
\label{subsec:rl}
In \ac{RL}, a \ac{MDP} is applied to model the synchronous interaction between an \emph{agent} and an \emph{environment}. An \ac{MDP} is defined by a 5-tuple $\left< \States, \Actions, \Rewardfunc, \Transition, \gamma \right>$, which comprises of a set of states $\States$, a set of actions $\Actions$, a reward function $\Rewardfunc$, a transition function $\Transition$ and the discount rate $\gamma \in [0, 1)$. Given a state $\state{t} \in \States$ at time $t$, an agent carries out an action $\action{t} \in \Actions(\state{t})$ from a set of available actions on  $\state{t}$. Upon taking action, the agent receives a reward \reward{t+1} and reaches the next state \state{t+1}. Hence, \reward{t+1} corresponds to the reward of taking action \action{t} on \state{t} and reaching \state{t+1}.\footnote{We follow the terminology used in~\cite{sutton2018reinforcement}.} The probability of reaching $\state{t+1}$ and receiving $\reward{t+1}$ by taking action $\action{t}$ in a given $\state{t}$ assigned by $\Transition$. This synchronous interaction between an agent and environment induces a \emph{trajectory}. A trajectory is a sequence of states, actions and rewards. The \emph{discounted return} of the $t^{th}$ item in a trajectory is defined as
\begin{equation}
    G_t = \reward{t+1} + \gamma \reward{t+2} +\gamma^2 \reward{t+3} + \dots = \sum_{k=0} \gamma^k \reward{t+k+1},
    \label{eq:return_func}
\end{equation}
where the discount rate $0 \leq \gamma < 1$ determines the present value of future rewards. The goal of an \ac{RL} agent is to select actions in a fashion that maximizes the discounted return~\cite{sutton2018reinforcement}.
\subsubsection{Acting optimally.}
A policy \policy prescribes which action to take in a given state. An optimal policy $\policy^*$ hence prescribes %a \ac{RL} agent 
optimal actions on any states so that $G_t$ is maximized. To obtain $\policy^*$, value functions are often used. 
Let $V_\policy: \States \mapsto \RealNumbers$ be the state-value function for a policy $\policy$. $V_\policy( \textbf{s})$ is defined as
\begin{align}
    V_\policy( \textbf{s}) &= \mathbb{E}_\policy \left[ G_t \mid \state{t}=\textbf{s} \right],\\
      &= \mathbb{E}_\policy \left[ \reward{t+1} + \gamma G_{t+1} \mid \state{t}=\textbf{s} \right]\\
                          &= \mathbb{E}_\policy \left[ \sum \nolimits_{k=0}^{\infty} \gamma^{k} \reward{t+k+1} \mid \state{t}= \textbf{s}\right],
    \label{eq:v_func}
\end{align}
where $\mathbb{E}_\policy \left[ \cdot\right]$ stands for the expected value of a random variable given that \policy is followed. Similarly,The action-value function $Q_\policy: \States \times \Actions  \mapsto \RealNumbers$  of a policy $\policy$ for a state $\textbf{s}$ and an action $\textbf{a}$ is defined as
\begin{align}
    Q_\policy( \textbf{s},\textbf{a}) &= \mathbb{E}_\policy \left[ G_t \mid \state{t}=\textbf{s}, \action{t}=\textbf{a}\right],\\
    &= \mathbb{E}_\policy \left[ \reward{t+1} + \gamma G_{t+1} \mid \state{t}=\textbf{s}, \action{t}=\textbf{a} \right]\\
     &= \mathbb{E}_\policy \left[ \sum \nolimits_{k=0}^{\infty} \gamma^{k} \reward{t+k+1} \mid \state{t}=\textbf{s}, \action{t}=\textbf{a}\right].
    \label{eq:q_func}
\end{align}
Using the Bellman equation, optimal value functions can be obtained recursively:
% % (1)
% \begin{equation}
% V_\policy ( \textbf{s}) =\sum_{\textbf{a}} \policy(\textbf{a} | \textbf{s})
%     \sum_{\textbf{s}', \reward{}} \Prob (\textbf{s}', \reward{} \;|\; \textbf{s}, \textbf{a}) \big[ \reward{} + \gamma V_\policy ( \textbf{s}' ) \big] \\
% \label{eq:bellman_v_func}    
% \end{equation}
% %
% \begin{equation}
% Q_\policy ( \textbf{s},\textbf{a}) = \sum_{\textbf{s}', \reward{} } \Prob (\textbf{s}', \reward{}\;|\; \textbf{s}, \textbf{a}) \big[ \reward{} + \gamma \max_{\textbf{a}' \in \Actions(\textbf{s}')}Q_\policy ( \textbf{s}',\textbf{a}') \big],
% \label{eq:bellman_q_func}
% \end{equation}
% %
% where $\policy(\textbf{a} | \textbf{s})$, $\Prob (\textbf{s}', \reward{} \;|\; \textbf{s}, \textbf{a})$ and $\textbf{a} \in \Actions(\textbf{s})$ and $\textbf{s}' \in \States(\textbf{s})$ denote the probability of taking action \textbf{a} on a given state \textbf{s}, the probability of  receiving \reward{} and transitioning $\textbf{s}'$ given that \textbf{a} is taken on \textbf{s}, and an action available on state \textbf{s} and a next reachable state from on state \textbf{s}, respectively.
% (1)
\begin{align}
    V_* ( \textbf{s})&= 
    \max_{\textbf{a} \in \Actions(\textbf{s})} \mathbb{E} \left[ \reward{t+1} + \gamma V_* ( \state{t+1}) \mid \state{t}=\textbf{s},\action{t}=\textbf{a} \right];
    %\\
    % &= \max_{\textbf{a} \in \Actions(\textbf{s})} \sum_{\textbf{s}', \reward{}} \Prob (\textbf{s}', \reward{} \;|\; \textbf{s}, \textbf{a}) \big[ \reward{} + \gamma V_* ( \textbf{s}' ) \big].
    \label{eq:bellman_v_func}    
\end{align}
\begin{align}
    Q_* (\state{},\action{})&= \mathbb{E} \left[ \reward{t+1} + \gamma \max_{\action{}' \in \Actions(\state{t+1})} Q_*(\state{t+1},\action{}') \mid \state{t}=\textbf{s},\action{t}=\textbf{a} \right].
    % \\
    % &=\sum_{\textbf{s}', \reward{}} \Prob (\textbf{s}', \reward{} \;|\; \textbf{s}, \textbf{a}) \big[ \reward{} + \max_{\action{}' \in \Actions(\state{t+1})} Q_* (\state{t+1},\action{}') \big].
    \label{eq:bellman_q_func}    
\end{align}
~\Cref{eq:bellman_q_func} establishes the relationship between the optimal value of a state-action pair and its successor state-action pairs~\cite{sutton2018reinforcement}. 
% The optimal the action-value function is defined as 
% $Q_* ( \textbf{s},\textbf{a}) = \max\limits_{\policy} \,  Q_\policy ( \textbf{s},\textbf{a} )$.
% \begin{equation}
% V_* ( \textbf{s}) = \max\limits_{\policy\in \AllPolicies} \,  V_\policy ( \textbf{s} )
% \label{eq:opt_v_func}
% \end{equation}
% \begin{equation}
% Q_* ( \textbf{s},\textbf{a}) = \max\limits_{\policy\in \AllPolicies} \,  Q_\policy ( \textbf{s},\textbf{a} )
% \label{eq:opt_q_func}
% \end{equation}
An optimal policy $\policy_*$ can be readily obtained if $Q_*$ is known, since any policy that is greedy with respect to $Q_*$ is optimal.\footnote{We refer Section 3.8 in \cite{sutton2018reinforcement} for the derivation of $Q_*$ by using the Bellman equation.}
% For instance, an optimal policy $\policy_*$ can be obtained directly from $Q_* (\textbf{s},\textbf{a})$\cite{franccois2018introduction}:
% \begin{equation}
%         \policy_*(\textbf{s}) = \operatorname*{argmax}_{\textbf{a} \in \Actions(\textbf{s})} Q_*(\textbf{s},\textbf{a}).
% \end{equation}
%
\subsubsection{Learning to act optimally.}
% To obtain $\policy_*$, it is common practice to iteratively approximate an optimal value function. For instance, 
$Q_*$ can be approximated by using the iterative Q-learning update:
\begin{equation}
Q_{i+1} (\textbf{s}_t,\textbf{a}_t) \leftarrow Q_{i} (\textbf{s}_t,\textbf{a}_t) + \alpha \big[ \reward{t+1} + \gamma \max_{\textbf{a} \in \Actions(\textbf{s}_{t+1})}Q_{i}( \textbf{s}_{t+1},\textbf{a}) -Q_{i} (\textbf{s}_t,\textbf{a}_t)\big],
    \label{eq:q_iterative_update}
\end{equation}
where the learning rate is denoted with $\alpha \in (0,1]$. Through using the iterative update defined in \Cref{eq:q_iterative_update}, $Q_i$ converges to $Q_*$ as $i \rightarrow \infty$~\cite{sutton2018reinforcement}. However, iteratively approximating exact optimal values is often computationally infeasible. For instance, solving the Bellman equation for $Q_*$ for the game of backgammon (with about $10^{20}$ states) would require years of computation even on currently available computers~\cite{sutton2018reinforcement}. In practice, a neural network parameterized with $\Theta$ (known as a Q-network) is commonly applied to estimate the optimal action-value function, $Q(s,a; \Theta) \approx Q_*(s,a)$~\cite{mnih2015human}. This approach is known as deep Q-learning and allows to effectively apply reinforcement learning %incorporate such information 
even in problems with large state-action spaces~\cite{sutton2018reinforcement,mnih2015human}. 

\input{refinementOperator}

\section{Methodology}
\label{section:methodology}
\subsubsection{Intuition.} 
\label{subsubsec:intuition}
% A concise summary of the problem in CEL
Devising a suitable heuristic function is crucial in \ac{CEL}~\cite{lehmann2011class}. The search of $\hypotheses$ is steered by optimizing a heuristic $\heuristic: \StateSpace \times \StateSpace \mapsto \RealNumbers$ signals whether refining the input state assists to find $H\in \hypotheses$ as elucidated in~\Cref{subsec:cel_in_dl}. ~\Cref{eq:ocel,eq:celoe} show that state-of-the-art \ac{CEL} models rely on heuristic functions that determine promises without any consideration for future states. More specifically, $\heuristic(s_i,s_j)$ is computed without incorporating any information pertaining to $\{s_k | s_k \in \StateSpace \wedge s_k \preceq s_j \}$. In the \ac{RL} framework,  this is analogous to setting $\gamma=0$ in~\Cref{eq:return_func}, i.e., to setting the present value of future rewards to 0. This implies that heuristic functions of state-of-the-art \ac{CEL} models correspond to \textbf{myopic policies}, whose only concern is to maximize immediate rewards (see Section 3.3 in~\cite{sutton2018reinforcement}, or Section 3 in~\cite{chan2009stochastic}). Playing a chess under a myopic policy corresponds to ignoring any future configurations of the board while selecting one's move.

% A concise summary of our proposal to tackle CEL
We leverage the deep Q-learning framework to incorporate information pertaining to $\{s_k \mid s_k \in \StateSpace \wedge s_k \preceq s_j \}$ while determining the heuristic value of the transition from a state $s_j$ to a state $s_i$. We represent \ac{RL} states $\textbf{s}_i$ by means of a vector in $\RealNumbers^{ |\Retrievalfunc(c_i)| \times \embdim}$ obtained by embedding the elements of $\Retrievalfunc(c_i)$ in $\RealNumbers^d$. Ergo, the root \ac{RL} state $\textbf{s}_\top$ captures all relevant information pertaining to any states by means of embeddings of $\Retrievalfunc(\top)$. Taking an action $\textbf{a}_j$ on $\textbf{s}_i$ implies refining $s_i$ in the refinement tree and reaching a state $s_j$. Upon taking an action $\textbf{a}_j$, \approach  deterministically reaches a \ac{RL} state
$\textbf{s}_j$ and receives a reward, i.e., $\Prob( \textbf{s}_j, \textbf{r} \; | \; \textbf{s}_i, \textbf{a}_j) = 1$. 

The number of possible actions $|\Actions(\textbf{s}_i)|$ on a given RL state $\textbf{s}_i$ corresponds to $|\rho(s_i)|$. Hence, the number of actions per state is not fixed in this environment. This entails that the standard deep Q-network loss function can not be directly applied~\cite{mnih2015human}. To mitigate this issue, we adapt the state-state value function $Q(\textbf{s}_i, \textbf{s}_j)$. Edwards et al.~\cite{edwards2020estimating} showed that the state-state Q function naturally \emph{avoid redundant actions}. As the number of redundant actions increases, using the state-state Q function results in more favorable results than using the state-action Q function. With these considerations, we define the following loss function
\begin{equation}
        \mathcal{L}( \Theta) =  \underset{ \textbf{s}_i, \textbf{s}_j, \textbf{e}_+,\textbf{e}_- \sim \mathcal{D}}{\mathbb E} \Bigg[ \Big(
        \textbf{r} + \gamma \operatorname*{max}_{\textbf{s}_x \preceq \textbf{s}_k} Q(\textbf{s}_k,\textbf{s}_x \mid \textbf{e}_+,\textbf{e}_-  ) 
        \; - \; \heuristic_{\approach} (\textbf{s}_i, \textbf{s}_j, \textbf{e}_+,\textbf{e}_- )
        \Big)^2 \Bigg].
        \label{equation:lossfunc}
\end{equation}
$\theta$, $e_+ , e_-$, $\textbf{r}$, $\mathcal{D}$ 
and $\operatorname*{max}_{\textbf{s}_x \preceq \textbf{s}_k} Q(\textbf{s}_k,\textbf{s}_x | \textbf{e}_+,\textbf{e}_-)$ denote parameters of \approach, embeddings of $E^+$, $E^-$, the experience replay, a reward of reaching $\textbf{s}_j$ from $\textbf{s}_i$ w.r.t. $E^+$ and  $E^-$, and a maximum reachable Q-value from $\textbf{s}_j$, respectively.
\subsubsection{Approach.}
\label{subsubsec:approach}
To minimize~\Cref{equation:lossfunc}, we propose~\approach---a fully connected convolutional network parameterized with $\Theta = [\omega,\textbf{W},\textbf{H},\textbf{b}_1,\textbf{b}_2]$. More specifically, \approach is defined as 
\begin{equation}
    \heuristic_{\approach}( \textbf{x})  = f\Big( \vect(f \big[ \Psi( \textbf{x} ) \ast \omega ) \big]  \cdot \mathbf{W} + \mathbf{b}_1 \Big) \cdot \mathbf{H} + \mathbf{b}_2,
    \label{eq:drill_convolution}
\end{equation}
where \textbf{x} represents $[\textbf{s}_i, \textbf{s}_j, \textbf{e}_+, \textbf{e}_- ]$: $\textbf{s}_i \in \RealNumbers^{|\Retrievalfunc(c_i)| \times \embdim}$, $\textbf{s}_j \in \RealNumbers^{ |\Retrievalfunc(c_j)| \times \embdim}$, $\textbf{e}_+ \in \RealNumbers^{ |E^+| \times \embdim}$, and $\textbf{e}_- \in \RealNumbers^{ |E^-| \times \embdim}$. $\Psi(\cdot)$ converts \textbf{x} into $\RealNumbers^{4 \times \embdim}$ by averaging the embeddings of its input. Hence, cumulated discounted future rewards are determined by considering the centroid of embeddings.\footnote{This is clearly not the only viable approach to achieve this goal. In future, we plan to use LSTM and Transformer architectures instead of averaging embeddings.} Moreover, $f(\cdot)$, $\vect(\cdot)$, $\ast$ and $\omega$
correspond the rectified linear unit function (ReLU), a flattening operation, the convolution operation, and kernels in the convolution operation. Two consecutive affine transformations are denoted with $(\mathbf{W},\mathbf{b}_1)$ and $(\mathbf{H},\mathbf{b}_2)$, respectively. 

To train \approach, we designed an unsupervised training scenario to automatically create learning problems in a fashion akin to~\cite{mnih2015human}.
\subsection{Unsupervised Training}
\label{subsec:unsupervised_training}
\subsubsection{Learning Problem Generation.} We first randomly generate $n$ goal states. To this end, we iteratively apply $\rho$ in a randomized depth-first search manner starting from $\top$, i.e., our random state generator select the next state to refine by randomly selecting states and refining them starting from $\rho(\top)$. During this process, states $s$ that satisfy the length constraint $\{s \, | \,  1 \leq |s| \leq maxlen \}$ are stored. Initially, we set $maxlen=5$. To avoid storing similar states, we performed this task $m$ times. Consequently, we obtain at most $n \times m$ states. For each stored state, we compute all positive and negative examples, i.e., $E^+ = \Retrievalfunc(c) $ and $E^-= \Individuals \setminus \Retrievalfunc(c)$, respectively. This operation often results in creating imbalanced $E^+$ and $E^-$,  with $|E^-| >> |E^+|$ being common. To alleviate imbalanced examples, we randomly undersample the largest set of examples so that $|E^+| = |E^-|$ and repeat this process $\kappa$ times. Consequently, we generated at most $n \times m \times \kappa$ learning problems. During the learning problem generation, we also considered introducing a size constraint at learning problem generation, $\{c \, | \,  1 \leq |c| \leq maxlen \wedge .1 |\Individuals| \leq |\Retrievalfunc(c)| \leq .3|\Individuals|\}$.

\subsubsection{Training Procedure.}
\Cref{alg:drill} elucidates the training procedure. We design this procedure in a fashion akin to~\cite{mnih2015human}. For the sake of brevity, we used $\textbf{s}_i$ to denote the vector representation of embeddings of $c_i$ and its counterpart $s_i$ in the refinement tree. \Cref{alg:drill} explicitly shows that at $15^{th}$ line, a target max. Q-value $y_i$ corresponds to the sum of an immediate reward $\textbf{r}_i$ and maximum Q-value obtained on remaining transactions $\mathcal{T}[i:]$. 
\begin{algorithm}[htb]
\caption{\approach with deep Q-learning training procedure}
\centering
\small
\label{alg:drill}
\begin{algorithmic}[1]
\State \textbf{Require}: Replay memory $\mathcal{D}$, \# of episodes M, \# of actions T, update constant U, reward function $\Rewardfunc$ and refinement operator $\rho$
\State Initialize $\Theta$ with Glorot initialization
\For{m $=1,M$} \Comment{Sequence of Episodes}
\State Initialise the search process $\textbf{s}_0 = \top$ and the transition storage $\mathcal{T}$
    \For {$i=0,T$} \Comment{Sequence of Actions}
        \State Refine the current state $\rho(\textbf{s}_i)$
    	\State Select a random state $\textbf{s}_j \in \rho(\textbf{s}_i)$ with probability $\epsilon$
    	\State otherwise select $\textbf{s}_j = \max_{ \textbf{s}_j \in \rho(\textbf{s}_i)} \heuristic_{\approach}([\textbf{s}_i,\textbf{s}_j, \textbf{e}_+,\textbf{e}_- ]); \Theta)$
    	\State Compute reward $\textbf{r}_i=\Rewardfunc(\textbf{s}_i , \textbf{s}_j)$
    	\State Store transition $\mathcal{T}[i]=[\textbf{s}_i, \textbf{s}_j,\textbf{e}_+,\textbf{e}_-,\textbf{r}_i]$
    	\State Set $\textbf{s}_{i+1}= \textbf{s}_j$
    \EndFor
    \State Reduce $\epsilon$ with a constant \Comment{Reduction of Exploration}
    \For {$i=0,T$} \Comment{Target Q-values}
        \State Select transition $[\textbf{s}_i,\textbf{s}_j, \textbf{e}_+,\textbf{e}_- , \textbf{r}_i]=\mathcal{T}[i]$
        \State Compute target value $y_i$ in $\mathcal{T}[i:]$ according to \Cref{eq:return_func}
        \State Store $(\textbf{s}_i,\textbf{s}_j, \textbf{e}_+,\textbf{e}_-, y_i)$ in $\mathcal{D}$
    \EndFor
    \If{m \% U == 0} \Comment{Updating Parameters}
        \State Sample random minibatches from $\mathcal{D}$
        \State Compute loss of minibatches w.r.t. \Cref{equation:lossfunc}
        \State Update $\Theta$ accordingly
    \EndIf
\EndFor
\end{algorithmic}
\end{algorithm}
\subsubsection{Construction of Rewards and Connection to CELOE.}
To compute rewards, we rely on 
$\heuristic_{\text{CELOE}}$ and defined the reward function as 
\begin{equation}
  \Rewardfunc(\textbf{s}_i,\textbf{s}_j)=\begin{cases}
                        maxreward, & \text{if $\text{F1-measure}(c_j)=1$}.\\
                        \heuristic_{\text{CELOE}}(c_i,c_j), & \text{otherwise}.
  \end{cases}
  \label{eq:reward_func_for_drill}
\end{equation}
\Cref{eq:reward_func_for_drill} shows that \approach is expected to steer the search towards shorter class expressions as $\heuristic_{\text{CELOE}}$ favors shorter class expressions (see \Cref{eq:celoe}). 

\section{Experiments}
\label{sec:Experiments}
\subsection{Datasets}
\label{subsec:dataset}
We used four benchmark datasets (Family, Carcinogenesis, Mutagenesis and Biopax) to evaluate \approach~\cite{lehmann2010concept,bin2016towards,fanizzi2008dl,fanizzi2019boosting}. An overview of the datasets is provided in~\Cref{tab:datasets}. 
We refer~\cite{bin2016towards,fanizzi2019boosting} for details pertaining to benchmark datasets.
\input{tables/dataset}
\subsection{Experimental Setup}
\label{subsec:experimental_setup}
We based our experimental setup on~\cite{buhmann2016dl}. Hence, we evaluated \approach by using the learning problems provided therein. In our experiments, we evaluate \approach in $\mathcal{ALC}$ for \ac{CEL}. However, \approach can be ported to all Description Logics. All experiments were carried out on Ubuntu 18.04 with $16$ GB RAM with Intel(R) Core(TM) i5-7300U CPU v4 @ 2.60GHz.
\paragraph{\textbf{Training}.} To train \approach, we first generate 10 learning problems for each benchmark datasets described in~\Cref{subsec:unsupervised_training}. During our experiments, we used the pretrained embeddings of input knowledge graphs provided by~\cite{demir2021convolutional,demir2021shallow}. We used ConEx embeddings of Family, Biopax, Mutagenesis and Shallom embeddings of Carcinogenesis. For more details about configurations of models, we refer project pages of~\cite{demir2021convolutional,demir2021shallow}.
For each generated learning problem, we trained \approach in an $\epsilon$-greedy fashion by using the the following configuration: ADAM optimizer with learning rate of .01, mini-batches of size 512, number of episodes set to 100, an epsilon decay of .01, a discounting factor $\gamma$ of .99, 32 input channels, and $(3 \times 3)$ kernels. The parameter selection was carried out based on~\cite{demir2021convolutional,mnih2015human}. The offline training on all benchmark datasets took 21 minutes on Biopax, 18 minutes on Family, 143 minutes on Mutagenesis and 119 on Carcinogenesis. Note that the offline training time is not relevant for the evaluation of \ac{CEL} as \ac{CEL} occurs online. %Moreover, the elapsed time during our offline training phase is infinitesimal compared to the state-of-the-art~\cite{Silver16Go}.
\paragraph{\textbf{Learning Problem Generation.}}
To perform an extensive comparison between approaches, we automatically generate  learning problems by using the procedure defined in~\Cref{subsec:unsupervised_training}. During our evaluation, we ensured that the set of learning problems used during training for \approach did not overlap with the set of learning problems used for evaluation.
\paragraph{\textbf{Stopping Criteria.}}
We used two standard stopping criteria for approaches. 
\begin{inparaenum}[(i)]
\item We set the maximum runtime to 3 seconds as Lehmann et al.~\cite{lehmann2010concept} showed that models often reach good solutions within 1.5 seconds. 
\item Approaches were configured to terminate if a goal state found.
\end{inparaenum}
\paragraph{\textbf{Evaluation Metrics.}}
\label{subsec:evaluation_metrics}
We compared approaches via F1-score, accuracy, the runtime and number of tested class expressions as similarly done in in~\cite{lehmann2010concept}. The F1-score and accuracy were used to measure the quality of the class expressions found w.r.t. positive and negative examples, while the runtime and the number of tested class expressions were used to measure the efficiency. 
\paragraph{\textbf{Implementation Details and Reproducibility.}}
\label{subsec:implementation_reproducability}
To ensure reproducibility of our results, We provide an open-source implementation of our approach, including training and evaluation scripts as well as pretrained models.\footnote{\url{https://github.com/dice-group/DRILL}}
\section{Results}

\Cref{table:results_small} reports results on standard learning problems provided within the DL-Learner framework~\cite{buhmann2016dl}. These results suggest that approaches yield similar performances in terms of F1-score and accuracy on benchmark datasets. However,~\approach outperforms all other approaches on all datasets w.r.t. its runtime. On all benchmark datasets, \approach requires at most \textbf{4 seconds} to yield competitive performance, while CELOE and ELTL require at most \textbf{21 seconds} and \textbf{18 seconds}, respectively. Hence, \approach is at least \textbf{2.7 times} more time-efficient than CELOE, OCEL and ELTL on all standard learning problems. 
During our experiments, we observed that (1) the number of tested class expressions in ELTL and (2) the F1-score of the best found class expression in OCEL are not reported.

\Cref{table:results_small} indicates that CELOE explores less number of states than other approaches. This may stem from (1) redundancy elimination and (2) expression simplification. (1) entails to query whether an expression already exists in the search tree. If yes, then this expression is not added into the search tree (see section 6.1 in~\cite{lehmann2007refinement}). (2) reduces long expressions into shorter ones, e.g., $\top \sqcup \texttt{Person}$ and $\forall r.\top$ into \texttt{Person} and $\top$, respectively. These modifications often lead to explore less number of states, i.e. reduce the memory requirement. However, they often introduce extra computations, i.e., increase runtime.

\input{tables/single_results}

% Our results on the Family dataset are shown in \Cref{table:family_results}. The table indicates that \approach is at least \textbf{4 times} more time-efficient than CELOE, OCEL and ELTL as \approach on average requires only \textbf{1.34 seconds} on the standard learning problems provided in the DL-Learner framework~\cite{lehmann2009dl}. Such distinct difference is not observed in terms of the effectiveness. 
% \Cref{table:family_results} show that CELOE explores less number of states than other approaches. This may stem from (1) redundancy elimination and (2) expression simplification. (1) entails to query whether an expression already exists in the search tree. If yes, then this expression is not added into the search tree (see section 6.1 in~\cite{lehmann2007refinement}). (2) reduces long expressions into shorter ones, e.g., $\top \sqcup \texttt{Person}$ and $\forall r.\top$ into \texttt{Person} and $\top$, respectively. These modifications may lead to explore less number of states, i.e. reduce the memory requirement. However, they often introduce extra computations.

\input{tables/family_results}
Note that we also evaluated DL-FOIL in our initial experiments. However, we observed that DL-FOIL often failed to terminate within 5 minutes. This may stem from the fact:
\begin{inparaenum}[(i)]
\item DL-FOIL does not use the elapsed time as a stopping criterion (see Section 4~\cite{fanizzi2008dl}). 
\item DL-FOIL requires to find an expression not involving $E^-$ to terminate (see Figure 1 in~\cite{fanizzi2008dl}).
\end{inparaenum}
Consequently, we could not include DL-FOIL in our experiments.

~\Cref{table:family_results_detail} show the details of our evaluation on 18 learning problems on the Family dataset in detail. These details suggest that \approach and CELOE often converge towards shorter concepts compared to OCEL and ELTL. This may stem from the fact that \approach indeed learned to assign low values for longer concepts as \approach is trained on rewards based on the heuristic function of CELOE.~\approach finds a goal state within a second in 12 out of 18 learning problems. During our experiments, we observed that approaches implemented in the DL-Learner framework do not terminate within the set maximum runtime. We delved into their implementations in the DL-Learner framework and observed that the maximum runtime criterion is not checked until refinements of a given class expression are obtained. 
During our experiments, we observed that there are only few learning problems per dataset. To perform more extensive comparisons between approaches, we automatically and randomly generate more learning problems.~\Cref{table:results_on_generated_lps} reports results on 370 learning problems generated automatically on benchmark datasets. These results confirm that \approach finds a goal state faster than all other approaches, while CELOE explores fewer expressions than all other approaches.
\input{tables/main_results}
\paragraph{\textbf{Statistical Hypothesis Testing.}}
We carried out a Wilcoxon signed-rank test to check whether our results are significant. Our null hypothesis was that the performances of \approach and CELOE come from the same distribution provided that a goal state is found. The alternative hypothesis was correspondingly that these results come from different distributions. To perform the Wilcoxon signed-rank test (two-sided), we used the differences of the runtimes of \approach and CELOE on benchmark datasets provided both approach found a goal node (F1-score 1.0). In Wilcoxon signed-rank test, we were able to reject the null hypothesis  with a p-value $< 1\%$. Ergo, the superior performance of \approach is statistically significant.
\paragraph{\textbf{Parameter Analysis and Optimization.}}
\approach achieves state-of-the-art performance on all datasets without requiring extensive parameter optimization. Throughout our experiments, \approach is trained with a fixed configuration: 32 input channels, (3x3) kernel and \approach has only 1.3 million parameters.
\subsubsection{Discussion}
Our results on all benchmark datasets suggest that \approach achieves a state-of-the-art performance w.r.t. the quality of the concepts it generates while outperforming the state of the art significantly w.r.t. its runtime. This improvement in performance is due to the following: 
\begin{inparaenum}[(i)]
\item deep Q-learning, 
\item the length-based refinement operator and 
\item the efficient computation of heuristic values.
\end{inparaenum}
Deep Q-learning endows \approach with the ability of considering future rewards while selecting the next state for the refinement. State-of-the-art approaches lack this ability. When using a myopic reward function, our refinement operator would lead to more refinements being considered than a subsumption-based refinement operator. However, using a length-based refinement operator allows \approach to detect redundant states. This ability results in reducing number of times the refinement operator applied during the search. % Additionally, our implementation of \approach is able to leverage parallel computation of heuristic values as neural network libraries (e.g., Pytorch) allow parallel computing with an ease. 
In our experimental setting, we observe that \approach is able to estimate promises of more than $10^3$ states less than a second on standard hardware. 
\section{Conclusion and Future Work}
\label{sec:conclusion}
In this work, we introduced \approach---a novel class expression learning approach that leverages a convolutional deep Q-network to accelerate the learning of concepts in \ALC.
% a class expression learning approach which uses a convolutional deep Q-learning model as heuristic function. 
To our knowledge, \approach is the first deep reinforcement model that is used to learn class expressions in $\mathcal{ALC}$. By virtue of using reinforcement learning, \approach is endowed with the capability of incorporating information pertaining to future rewards while determining the heuristic value of states. \ac{CEL} approaches lack this ability as they rely on myopic heuristic functions to guide their search through an infinite concept space. Our experiments show that \approach outperforms state-of-the-art approaches w.r.t. its runtime. Our statistical hypothesis test confirms the superior performance of \approach on all benchmark datasets.

In future work, we plan to investigate: 
\begin{inparaenum}[(i)]
\item learning embeddings during the training phase, 
\item using LSTM and Transformer architecture in \approach, 
\item constructing a reward function based on the gain function in DL-FOIL, and 
\item leveraging  reinforcement learning at generating difficult learning problems.
\end{inparaenum}

\bibliographystyle{splncs04}
\bibliography{references}
\end{document}

%% file: acronyms.tex
\begin{acronym}[UML]
	\acro{AI}{Artificial Intelligence}
	\acro{BFS}{Breadth-First-Search}
	\acro{BoW}{Bag-of-Words}
	\acro{CEL}{Class Expression Learning}
	\acro{CL}{Concept Learning}
	
	\acro{DL}{Deep Learning}

	\acro{ER}{Entity Resolution}
	\acro{EM}{Expectation Maximization}
	\acro{EBMT}{Example-Based Machine Translation}
	\acro{EBNF}{Extended Backus--Naur Form}
	\acro{EL}{Entity Linking}
	\acro{FAO}{Food and Agriculture Organization of the United Nations}
	\acro{GIS}{Geographic Information Systems}
	\acro{GHO}{Global Health Observatory}
	\acro{GRU}{Gated recurrent unit}
	\acro{HDI}{Human Development Index}
	\acro{ILP}{Inductive Logic Programming}

    \acro{KB}{Knowledge Base}
    \acro{KG}{Knowledge Graph}
    \acrodefplural{KG}{Knowledge Graphs}
    \acro{KGE}{Knowledge Graph Embeddings}
    \acro{KBSE}{Knowledge Base Semantic Embedding}
    \acro{KGC}{Knowledge Graph Completion}
	\acro{LR}  {Language Resource}
	\acro{LD}  {Linked Data}
	\acro{LLOD}  {Linguistic Linked Open Data}
	\acro{LIMES}{LInk discovery framework for MEtric Spaces}
	\acro{LS}  {Link Specifications}
	\acro{LDIF}{Linked Data Integration Framework}
	\acro{LGD} {LinkedGeoData}
	\acro{LOD} {Linked Open Data}
	\acro{LOV} {Linked Open Vocabularies}
	\acro{LSTM}{Long Short-Term Memories}
	\acro{MSE}{Mean Squared Error}
	\acro{MWE}{Multiword Expressions}
	\acro{MT}{Machine Translation}
	\acro{ML}{Machine Learning}
	\acro{MR}{Machine Reading}
	\acro{NIF}{Natural Language Processing Interchange Format}
	\acro{NIF4OGGD}{NLP Interchange Format for Open German Governmental Data}
	\acro{NLP}{Natural Language Processing}
	\acro{NER}{Named Entity Recognition}
	\acro{NMT}{Neural Machine Translation}
	\acro{NN}{Neural Network}
	\acrodefplural{NN}{Neural Networks}
	\acro{NLG}{Natural Language Generation}
	\acro{NED}{Named Entity Disambiguation}
	\acro{NERD}{Named Entity Recognition and Disambiguation}
	\acro{NL}{Natural Language}
	\acro{NIF}{NLP Interchange Format}
	\acro{NIF4OGGD}{NLP Interchange Format for Open German Governmental Data}
	\acro{NLP}{Natural Language Processing}
	\acro{NER}{Named Entity Recognition}
	\acro{NEL}{Named Entity Linking}
	\acro{NE}{Named Entity}
	\acro{NN}{Neural Network}
	\acro{NLI}{Natural Language Inference}
	\acro{OSM}{OpenStreetMap}
	\acro{OWL}{Web Ontology Language}
	\acro{OOV}{out-of-vocabulary}
	\acro{PFM}{Pseudo-F-Measures}
	\acro{PSO}{Particle-Swarm Optimization}
	\acro{PBSMT}{Phrase-Based Statistical Machine Translation}
	
	\acro{QA}{Question Answering}
	\acro{RDF}{Resource Description Framework}
	\acro{RBMT}{Rule-Based Machine Translation}
	\acro{RNN}{Recurrent Neural Network}
	\acro{ReLU}{rectified linear unit}
	\acro{RDFS}{RDF Schema}
	\acro{SPARQL}{SPARQL Protocol and RDF Query Language}
	\acro{SML}{Structured Machine Learning}
	\acro{SRL}{Statistical Relational Learning}
	\acro{SF}{surface forms}
    \acro{TBMT} {Transfer-Based Machine Translation}
	\acro{UML}{Unified Modeling Language}
	\acro{USL}{Ukrainian Sign Language}
	\acro{URI}{Uniform Resource Identifier}
	\acro{WHO}{World Health Organization}
	\acro{WKT}{Well-Known Text}
	\acro{W3C}{World Wide Web Consortium}
	\acro{WSD}{Word Sense Disambiguation}
	\acro{WMT}{Workshop on Machine Translation}
    \acro{XML}{Extensible Markup Language}
	\acro{YPLL}{Years of Potential Life Lost}
	\acro{AOS}{Agricultural Ontology Services}
	\acro{AGRIS}{Agricultural Science and Technology}
	\acro{API}{Application Programming Interface}
	\acro{AOS}{Agricultural Ontology Services}
	\acro{AGRIS}{Agricultural Science and Technology}
	\acro{API}{Application Programming Interface}
	\acro{A2KB}{Annotation to Knowledge Base}
	\acro{BPSO}{Binary Particle-Swarm Optimization}
	\acro{BPMLOD}{Best Practices for Multilingual Linked Open Data}
	\acro{BPSO}{Binary Particle-Swarm Optimization}
	\acro{BPMLOD}{Best Practices for Multilingual Linked Open Data}
	\acro{BFS}{Breadth-First-Search}
	\acro{BPE}{Byte Pair Encoding}
	\acro{BoW}{Bag-of-Words}
	\acro{CBD}{Concise Bounded Description}
	\acro{COG}{Content Oriented Guidelines}
	\acro{CSV}{Comma-Separated Values}
	\acro{CBMT}{Corpus-Based Machine Translation}
	\acro{CLIR}{Cross-Language Information Retrieval}
	\acro{DPSO}{Deterministic Particle-Swarm Optimization}
	\acro{DALY}{Disability Adjusted Life Year}

	\acro{ER}{Entity Resolution}
	\acro{EM}{Expectation Maximization}
	\acro{EBMT}{Example-Based Machine Translation}
	\acro{EBNF}{Extended Backus--Naur Form}
	\acro{EL}{Entity Linking}
	\acro{FAO}{Food and Agriculture Organization of the United Nations}
	\acro{GIS}{Geographic Information Systems}
	\acro{GHO}{Global Health Observatory}
	\acro{GRU}{Gated recurrent unit}
	\acro{HDI}{Human Development Index}
	\acro{ICT}{Information and communication technologies}
	\acro{IFRS}{International Financial Reporting Standards}
	\acro{ICD}{International Classification of Diseases}
	\acro{IT}{Information Technology}
	\acro{IRI}{International Resource Identifier}
    \acro{KB}{Knowledge Base}
    \acro{KG}{Knowledge Graph}
    \acro{KGE}{Knowledge Graph Embeddings}
    \acro{KBSE}{Knowledge Base Semantic Embedding}
	\acro{LR}  {Language Resource}
	\acro{LD}  {Linked Data}
	\acro{LLOD}  {Linguistic Linked Open Data}
	\acro{LIMES}{LInk discovery framework for MEtric Spaces}
	\acro{LS}  {Link Specifications}
	\acro{LDIF}{Linked Data Integration Framework}
	\acro{LGD} {LinkedGeoData}
	\acro{LOD} {Linked Open Data}
	\acro{LSTM}{Long Short-Term Memories}
	\acro{MSE}{Mean Squared Error}
	\acro{MWE}{Multiword Expressions}
	\acro{MT}{Machine Translation}
	\acro{ML}{Machine Learning}
	\acro{MR}{Machine Reading}
	\acro{MOS}{Manchester OWL Syntax}
	\acro{NIF}{Natural Language Processing Interchange Format}
	\acro{NIF4OGGD}{NLP Interchange Format for Open German Governmental Data}
	\acro{NLP}{Natural Language Processing}
	\acro{NER}{Named Entity Recognition}
	\acro{NMT}{Neural Machine Translation}
	\acro{NN}{Neural Network}
	\acro{NLG}{Natural Language Generation}
	\acro{NED}{Named Entity Disambiguation}
	\acro{NERD}{Named Entity Recognition and Disambiguation}
	\acro{NL}{Natural Language}
	\acro{NIF}{NLP Interchange Format}
	\acro{NIF4OGGD}{NLP Interchange Format for Open German Governmental Data}
	\acro{NLP}{Natural Language Processing}
	\acro{NER}{Named Entity Recognition}
	\acro{NEL}{Named Entity Linking}
	\acro{NE}{Named Entity}
	\acro{NN}{Neural Network}
	\acro{NLI}{Natural Language Inference}
	\acro{OSM}{OpenStreetMap}
	\acro{OWL}{Web Ontology Language}
	\acro{OOV}{out-of-vocabulary}
	\acro{PFM}{Pseudo-F-Measures}
	\acro{PSO}{Particle-Swarm Optimization}
	\acro{PBSMT}{Phrase-Based Statistical Machine Translation}
	
	\acro{QA}{Question Answering}
	\acro{RDF}{Resource Description Framework}
	\acro{RBMT}{Rule-Based Machine Translation}
	\acro{RNN}{Recurrent Neural Network}
	\acro{ReLU}{rectified linear unit}
	\acro{SKOS}{Simple Knowledge Organization System}
	\acro{SPARQL}{SPARQL Protocol and RDF Query Language}
	\acro{SRL}{Statistical Relational Learning}
	\acro{SWT}{Semantic Web Technologies}
	\acro{SW}{Semantic Web}
	\acro{SMT}{Statistical Machine Translation}
	\acro{SWMT}{Semantic Web Machine Translation}
	\acro{SKOS}{Simple Knowledge Organization System}
	\acro{SPARQL}{SPARQL Protocol and RDF Query Language}
	\acro{SRL}{Statistical Relational Learning}
	\acro{SF}{surface forms}
	\acro{SVM}{Support Vector Machines}
    \acro{TBMT} {Transfer-Based Machine Translation}
	\acro{UML}{Unified Modeling Language}
	\acro{USL}{Ukrainian Sign Language}
	\acro{URI}{Uniform Resource Identifier}
	\acro{WHO}{World Health Organization}
	\acro{WKT}{Well-Known Text}
	\acro{W3C}{World Wide Web Consortium}
	\acro{WSD}{Word Sense Disambiguation}
	\acro{WWW}{World Wide Web}
    \acro{XML}{Extensible Markup Language}
	\acro{YPLL}{Years of Potential Life Lost}
	\acro{AOS}{Agricultural Ontology Services}
	\acro{AGRIS}{Agricultural Science and Technology}
	\acro{API}{Application Programming Interface}
	\acro{BPSO}{Binary Particle-Swarm Optimization}
	\acro{BPMLOD}{Best Practices for Multilingual Linked Open Data}
	\acro{CBD}{Concise Bounded Description}
	\acro{COG}{Content Oriented Guidelines}
	\acro{CSV}{Comma-Separated Values}
	\acro{CBMT}{Corpus-Based Machine Translation}
	\acro{CLIR}{Cross-Language Information Retrieval}
	\acro{DPSO}{Deterministic Particle-Swarm Optimization}
	\acro{DALY}{Disability Adjusted Life Year}
	\acro{DRL}{Deep Reinforcement Learning}

	\acro{ER}{Entity Resolution}
	\acro{EM}{Expectation Maximization}
	\acro{EBMT}{Example-Based Machine Translation}
	\acro{EBNF}{Extended Backus--Naur Form}
	\acro{EL}{Entity Linking}
	\acro{FAO}{Food and Agriculture Organization of the United Nations}
	\acro{GIS}{Geographic Information Systems}
	\acro{GHO}{Global Health Observatory}
	\acro{HDI}{Human Development Index}
	\acro{ICT}{Information and communication technologies}
    \acro{KB}{Knowledge Base}
    \acro{KBSE}{Knowledge Base Semantic Embedding}
	\acro{LR}  {Language Resource}
	\acro{LD}  {Linked Data}
	\acro{LLOD}  {Linguistic Linked Open Data}
	\acro{LIMES}{LInk discovery framework for MEtric Spaces}
	\acro{LS}  {Link Specifications}
	\acro{LDIF}{Linked Data Integration Framework}
	\acro{LGD} {LinkedGeoData}
	\acro{LOD} {Linked Open Data}
	\acro{ML}{Machine Learning}
	\acro{MDP}{Markov Decision Process}
	\acro{MT}{Machine Translation}
	\acro{NIF}{Natural Language Processing Interchange Format}
	\acro{NIF4OGGD}{NLP Interchange Format for Open German Governmental Data}
	\acro{NLP}{Natural Language Processing}
	\acro{NER}{Named Entity Recognition}
	\acro{NMT}{Neural Machine Translation}
	\acro{NN}{Neural Network}
	\acro{NLG}{Natural Language Generation}
	\acro{NED}{Named Entity Disambiguation}
	\acro{NERD}{Named Entity Recognition and Disambiguation}
	\acro{NL}{Natural Language}
	\acro{OWL}{Web Ontology Language}
	\acro{QA}{Question Answering}
	\acro{RDF}{Resource Description Framework}
	\acro{RBMT}{Ru\-le-Ba\-sed Ma\-chi\-ne Trans\-la\-tion}
	\acro{REG}{Referring Expression Generation}
	\acro{SKOS}{Simple Knowledge Organization System}
	\acro{SPARQL}{SPARQL Protocol and RDF Query Language}
	\acro{SRL}{Statistical Relational Learning}
	\acro{SWT}{Semantic Web Technologies}
	\acro{SW}{Semantic Web}
	\acro{SMT}{Statistical Machine Translation}
	\acro{SWMT}{Semantic Web Machine Translation}
    \acro{TBMT} {Transfer-Based Machine Translation}
	\acro{VS}{Version Space}

	\acro{WHO}{World Health Organization}
	\acro{WKT}{Well-Known Text}
	\acro{W3C}{World Wide Web Consortium}
	\acro{WSD}{Word Sense Disambiguation}
    \acro{XML}{Extensible Markup Language}
	\acro{YPLL}{Years of Potential Life Lost}
	\acro{AOS}{Agricultural Ontology Services}
	\acro{AGRIS}{Agricultural Science and Technology}
	\acro{API}{Application Programming Interface}
	\acro{AOS}{Agricultural Ontology Services}
	\acro{AGRIS}{Agricultural Science and Technology}
	\acro{API}{Application Programming Interface}
	\acro{A2KB}{Annotation to Knowledge Base}
	\acro{BPSO}{Binary Particle-Swarm Optimization}
	\acro{BPMLOD}{Best Practices for Multilingual Linked Open Data}
	\acro{BPSO}{Binary Particle-Swarm Optimization}
	\acro{BPMLOD}{Best Practices for Multilingual Linked Open Data}
	\acro{BFS}{Breadth-First-Search}
	\acro{BPE}{Byte Pair Encoding}
	\acro{BoW}{Bag-of-Words}
	\acro{CBD}{Concise Bounded Description}
	\acro{COG}{Content Oriented Guidelines}
	\acro{CSV}{Comma-Separated Values}
	\acro{CBMT}{Corpus-Based Machine Translation}
	\acro{CLIR}{Cross-Language Information Retrieval}
	\acro{DPSO}{Deterministic Particle-Swarm Optimization}
	\acro{DALY}{Disability Adjusted Life Year}

	\acro{ER}{Entity Resolution}
	\acro{EM}{Expectation Maximization}
	\acro{EBMT}{Example-Based Machine Translation}
	\acro{EBNF}{Extended Backus--Naur Form}
	\acro{EL}{Entity Linking}
	\acro{FAO}{Food and Agriculture Organization of the United Nations}
	\acro{GIS}{Geographic Information Systems}
	\acro{GHO}{Global Health Observatory}
	\acro{GRU}{Gated recurrent unit}
	\acro{HDI}{Human Development Index}
	\acro{ICT}{Information and communication technologies}
	\acro{IFRS}{International Financial Reporting Standards}
	\acro{ICD}{International Classification of Diseases}
	\acro{IT}{Information Technology}
    \acro{KB}{Knowledge Base}
    \acro{KG}{Knowledge Graph}
    \acro{KGE}{Knowledge Graph Embeddings}
    \acro{KBSE}{Knowledge Base Semantic Embedding}
	\acro{LR}  {Language Resource}
	\acro{LD}  {Linked Data}
	\acro{LLOD}  {Linguistic Linked Open Data}
	\acro{LIMES}{LInk discovery framework for MEtric Spaces}
	\acro{LS}  {Link Specifications}
	\acro{LDIF}{Linked Data Integration Framework}
	\acro{LGD} {LinkedGeoData}
	\acro{LOD} {Linked Open Data}
	\acro{LSTM}{Long Short-Term Memories}
	\acro{MSE}{Mean Squared Error}
	\acro{MWE}{Multiword Expressions}
	\acro{MT}{Machine Translation}
	\acro{ML}{Machine Learning}
	\acro{MR}{Machine Reading}
	\acro{NIF}{Natural Language Processing Interchange Format}
	\acro{NIF4OGGD}{NLP Interchange Format for Open German Governmental Data}
	\acro{NLP}{Natural Language Processing}
	\acro{NER}{Named Entity Recognition}
	\acro{NMT}{Neural Machine Translation}
	\acro{NN}{Neural Network}
	\acro{NLG}{Natural Language Generation}
	\acro{NED}{Named Entity Disambiguation}
	\acro{NERD}{Named Entity Recognition and Disambiguation}
	\acro{NL}{Natural Language}
	\acro{NIF}{NLP Interchange Format}
	\acro{NIF4OGGD}{NLP Interchange Format for Open German Governmental Data}
	\acro{NLP}{Natural Language Processing}
	\acro{NER}{Named Entity Recognition}
	\acro{NEL}{Named Entity Linking}
	\acro{NE}{Named Entity}
	\acro{NN}{Neural Network}
	\acro{NLI}{Natural Language Inference}
	\acro{OSM}{OpenStreetMap}
	\acro{OWL}{Web Ontology Language}
	\acro{OOV}{out-of-vocabulary}
	\acro{PFM}{Pseudo-F-Measures}
	\acro{PSO}{Particle-Swarm Optimization}
	\acro{PBSMT}{Phrase-Based Statistical Machine Translation}
	
	\acro{QA}{Question Answering}
	\acro{RL}{Reinforcement Learning}
	\acro{RDF}{Resource Description Framework}
	\acro{RNN}{Recurrent Neural Network}
	\acro{ReLU}{rectified linear unit}
	
	\acro{SKOS}{Simple Knowledge Organization System}
	\acro{SPARQL}{SPARQL Protocol and RDF Query Language}
	\acro{SRL}{Statistical Relational Learning}
	\acro{SWT}{Semantic Web Technologies}
	\acro{SW}{Semantic Web}
	\acro{SMT}{Statistical Machine Translation}
	\acro{SWMT}{Semantic Web Machine Translation}
	\acro{SKOS}{Simple Knowledge Organization System}
	\acro{SPARQL}{SPARQL Protocol and RDF Query Language}
	\acro{SRL}{Statistical Relational Learning}
	\acro{SF}{surface forms}
	\acro{SVM}{Support Vector Machines}
    \acro{TBMT} {Transfer-Based Machine Translation}
	\acro{UML}{Unified Modeling Language}
	\acro{USL}{Ukrainian Sign Language}
	\acro{URI}{Uniform Resource Identifier}
	\acro{WHO}{World Health Organization}
	\acro{WKT}{Well-Known Text}
	\acro{W3C}{World Wide Web Consortium}
	\acro{WSD}{Word Sense Disambiguation}
    \acro{XML}{Extensible Markup Language}
	\acro{YPLL}{Years of Potential Life Lost}
\end{acronym}  

%% file: notation.tex
\newcommand{\ABox}{\ensuremath{Abox}\xspace}
\newcommand{\TBox}{\ensuremath{Tbox}\xspace}
\newcommand{\Classes}{\ensuremath{C}\xspace}
\newcommand{\Individuals}{\ensuremath{I}\xspace}

\newcommand{\hypotheses}{\ensuremath{\mathcal{H}}\xspace}
\newcommand{\heuristic}{\ensuremath{\phi}\xspace}
\newcommand{\StateSpace}{\ensuremath{\mathcal{S}}\xspace}

\newcommand{\Retrievalfunc}{\ensuremath{\mathcal{R}}}

\newcommand{\ALC}{\ensuremath{\mathcal{ALC}}\xspace}
\newcommand{\lang}{\ensuremath{\mathscr{L}}\xspace}

\newcommand{\States}{\ensuremath{\textbf{S}}\xspace}
\newcommand{\Actions}{\ensuremath{\textbf{A}}\xspace}
\newcommand{\Rewardfunc}{\ensuremath{\textbf{R}}\xspace}
\newcommand{\Transition}{\ensuremath{\textbf{T}}\xspace}
\newcommand{\Prob}{\ensuremath{\mathbb{P}}\xspace}

\newcommand{\policy}{\ensuremath{\pi}\xspace}

\newcommand{\state}[1]{\ensuremath{\mathbf{s}_{#1}}}
\newcommand{\action}[1]{\ensuremath{\mathbf{a}_{#1}}}
\newcommand{\reward}[1]{\ensuremath{R_{#1}}}

\newcommand{\approach}{\textsc{Drill}\xspace}
\newcommand{\kg}{\ensuremath{\mathcal{G}}\xspace}

\newcommand{\embdim}{\ensuremath{d}}

\newcommand{\RealNumbers}{\ensuremath{\mathbb{R}}\xspace}
\newcommand{\NaturalNumber}{\ensuremath{\mathbb{N}}\xspace}
\DeclareMathOperator{\vect}{vec}

%% file: Images/SeachSpace.tikz
\begin{tikzpicture}[scale=1,baseline,thick]
  %% nodes or vertices
  \GraphInit[vstyle=Classic]
  \tikzset{vertex/.style ={draw=black,shape=rectangle,fill=white,minimum size=13pt,drop shadow}}
   \node at (2.5,4)[vertex] (thing) {$\rho(\top)$};      
   \node at (0,3)[vertex] (person) {$\rho(\text{Person})$}; 
   \node at (1.5,3) (subl) {\dots};
   \node at (2.5,3)[vertex] (place) {Place};
   \node at (3.5,3) (subl) {\dots};
   \node at (5,3)[vertex] (organisation) {Organisation};
  \node at (0,2)[vertex] (personthatworks) {$\rho(\text{Person}\sqcap\exists \text{worksIn.}\top$)};
  \node at (0,1)[vertex] (parentOfLeafs) {$\rho(\text{Person}\sqcap\exists\text{worksIn.University}$};
  \node at (2.9,1) (subl) {\dots};
  \node at (5.5,1)[vertex] (personAndAtLeastLikesLiterature) {Person $\sqcap \, \exists$ worksIn.Hospital};
  \node at (0,0)[vertex] (leaf1) {$\text{Person} \sqcap \exists \text{worksIn.MathematicsDept}$};
  \node at (6,0)[vertex] (leaf2) { Person $\sqcap \, \exists$ worksIn.PhysicsDept};
  %% edges
  \tikzstyle{EdgeStyle}=[->,>=stealth,thick]
  \Edge (parentOfLeafs)(leaf1) \Edge (parentOfLeafs)(leaf2)
  \Edge (personthatworks)(parentOfLeafs) \Edge (personthatworks)(personAndAtLeastLikesLiterature)
  \Edge (person)(personthatworks) 
  \Edge (thing)(place) 
  \Edge (thing)(organisation) 
  \Edge (thing)(person) 
\end{tikzpicture}

%% file: refinementOperator.tex
\section{Refinement Operator}
Let $N_C$ be a set of finite named concepts and let $R$ be a finite set of roles. We write $\top$ to denote the top concept and $\bot$ to denote the bottom concept.  We set $N_C^+ = N_C \cup \{\top, \bot \}$. We define $S$ to be the set of all $\mathcal{ALC}$ concepts built upon $N_C$ and $R$. Formally this the following:
 \begin{itemize}
     \item $N_C^+ \subset S$;
     \item If $C \in S$ and $D \in S$, then $(C \sqcap D) \in S$
     \item If $C \in S$ and $D \in S$, then $(C \sqcup D) \in S$
     \item If $C \in S$ and $r \in R$, then $\exists r.C \in S$
     \item If $C \in S$ and $r \in R$, then $\forall r.C \in S$
     \item If $C \in S$ then $(\neg C) \in S$
 \end{itemize}

Let $X, Y \in S$  and $r \in R$. We define the length of a concept $C \in S$ (denoted $|C|$) as follows:
\begin{itemize}
    \item If $C \in N_C^+$, then $|C| = 1$;
    \item If $C = X \sqcup Y$ or  $C = X \sqcap Y$, then $|C| = |X| + |Y| + 1$;
    \item If $C = \exists r.X$ or $C = \exists r.X$, then $|C| = |X| + 2$; 
    \item If $C = \neg X$, then $|C| = |X| + 1$. 
\end{itemize}
The length of a concept is an ordering over the set $S$. We define the operator $\rho$ over $(S, |\cdot|)$ as follows:
\begin{equation}
\rho(C) = \begin{cases}
\{\exists r.C, \forall r.C, C \sqcap \top, C \sqcup \top, \neg C, C\} & \mbox{for any $C$} \\
\{\exists r.\rho(X)\} &  \mbox{if $C = \exists r.X$}\\
\{\forall r.\rho(X)\} &  \mbox{if $C = \forall r.X$}\\
\{\neg \rho(X)\} &  \mbox{if $C = \neg X$}\\
\{\rho(X) \sqcup \rho(Y)\} &  \mbox{if $C = X \sqcup Y$}\\
\{\rho(X) \sqcap \rho(Y)\} &  \mbox{if $C = X \sqcap Y$}\\
N_C^+ &  \mbox{if $C = \top$}\\
\end{cases}
\end{equation}
In the following, we will assume that $\rho$ begins the refinement process from $\top$.

\begin{theorem} $\rho$ is an upward refinement operator over $(S, |\cdot|)$ 
\end{theorem}
\begin{proof}
We need to prove that $|\rho(C)| \geq |C|$. This is a direct consequence of the construction of $C$. For each possible refinement computed $\rho$ and by virtue of the definition of $|\cdot|$, it is easy to see that $\rho$ either preserves the length of a concept (e.g., if $C \in N_C^+$) or leads to longer concepts (e.g., for some refinements in the first line of the specification of $C$). $\square$
\end{proof}
In the following, we will use the following convention: the operator $\rho^*$ will stand for the transitive closure of $\rho$. Ergo, if $D$ can be generated from $C$ by $\rho$ through manifold application, we will simply write $D \in \rho(C)$.  

Every $\mathcal{ALC}$ concept $C$ can be represented as a tree. This is a direct consequence of the definition of \emph{ALC} concepts. For example, all concepts $C \in N_C^+$ can be represented by trees with exactly one node and not edges. If $C \notin N_C^+$, then $C$ can only take one of five forms: 

%%Add TikZ representations here
Two observations are important here. First, there can be several tree representations for any concept $C$. We 
define the height $h(C)$ by choosing the smallest height $h_{\min}$ over all possible tree representations of $C$ and setting the height of $C$, denoted $h(C)$, to exactly $h_{\min}$. %If several of these representations exist, we select any of these representations (e.g., the representation for which a globally fixed perfect hash function returns the smallest value). We write  $t(C)$ to denote the tree which represents the concept $C$.  

The second observation is that only elements of $N_C^+$ have height 0, which is a direct consequence of the construction of complex concepts in $\ALC{C}$: If the concept $C \notin N_C^+$, then it must have a length of at least 2 as its tree representation must hence contain at least 2 nodes. By virtue of the definition of the height of a tree, a tree with two nodes must have a height of at least 1.
%Now by induction over the length \emph{ALC} concepts, if $X$ and $Y$ can be represented as trees, so can $C$ as $|X| < |C|$ and $|Y| < |C|$.

\begin{lemma}
$\rho$ can generate any concept of height 0.
\end{lemma}
\begin{proof}
Note that only elements of $N_C^+$ have height 0. By virtue of the last line of its specification, $\rho$ can generate all elements of $N_C^+$ and thus all elements of height 0. $\square$
\end{proof}

\begin{lemma}
Assume $n \geq 0$. If $\rho$ can generate all concepts with a height $h \leq n$,  then it can generate all concepts of height $n+1$. 
\end{lemma}
\begin{proof}
A concept $C$ of height $n+1$ can be written in one of five ways:
\begin{enumerate}
    \item $C = \exists r.X$
    \item $C = \forall r.X$
    \item $C = \neg X$
    \item $C = X \sqcup Y$
    \item $C = X \sqcap Y$
\end{enumerate}
Let us begin by considering the first three possibilities. By definition of the height, $h(C) > h(X)$. Hence, by virtue of our assumption, $\rho$ can generate $X$. Now the first generation rule of $\rho$ states that $\exists r.X$, $\forall r.X$ and $\neg X$ belong to the possible refinements of $X$. Hence, $\rho$ can generate $C$ for these three cases.

We are left with two possibilities: $C = X \sqcup Y$ or $C = X \sqcap Y$. In both cases, we can assume that $|X| \geq |Y|$ without loss of generality. We prove that $\rho$ can generate $C = X \sqcap Y$. The other possibility can be addressed analogously. By the definition of height of a tree, $h(X) = n$. If it were not the case, then $h(Y) < n$ would also hold, leading to $h(X) < n+1$, which would contradict the assumption on the height of $X$ made in the premise of this proof. Now, given that $h(X) = n$, it can be generated by $\rho$ by virtue of the assumption of this lemma. Similarly, $h(Y) \leq n$ also means that it can be generated by $\rho$ by virtue of the assumption of this lemma. Remember that all concepts are generated from $\top$. We can now show that there must be a refinement path that leads to $C$. First, given that $\rho$ can generate $X$, there is a refinement path from $\top$ to $X$, i.e., $X \in \rho^*(\top)$. Consequently, $(X \sqcap \top) \in \rho^*(\top)$ by virtue of the first rule in the specification of $\rho$. $\rho(X  \sqcap \top) = \rho(X)  \sqcap \rho(\top)$. As $X \in \rho(X)$, we solely need to show that $Y \in \rho(\top)$. This is a direct consequence of $h(Y) \leq n$ as we assumed that every concept of height $n$ can be generated by $\rho$ and every concept that can be generated by $\rho$ is an element of $\rho^*(\top)$. $\square$ 
\end{proof}

\begin{theorem}
Given a set of named concepts $N_C$, $\rho$ can generate all $\mathcal{ALC}$ concepts based on $N_C$.
\end{theorem}
%add missing lemmas
%define \rho^*
%examples
%tikz images
\begin{proof}
This is a direct consequence of the two lemmas above.
\end{proof}
It is obvious that $\rho$ is \emph{redundant}, i.e., there can be more than one sequence of refinement from $\top$ to a concept $C$. The operator is \emph{finite} as $|\rho(C)| < \infty$ for any $\mathcal{ALC}$ concepts $C$ built upon a finite set of named concepts $N_C$ and a finite set of roles. We mostly exploit these two characteristics during the implementation of the refinement operator.

%It is easy to see that $\rho$ is redundant, 

%% file: tables/dataset.tex
\begin{table}
\centering
\small
\caption{Overview of benchmark datasets.}
    \begin{tabular}{@{}lccccc@{}}
        \toprule
        Dataset & \#instances & \#concepts & \#obj. properties & \#data properties & DL language \\
        \midrule
        Family & 202 & 18 & 4 & 0 & $\mathcal{ALC}$\\
        % Forte  & 86 & 4 & 3 & 0 & $\mathcal{ALC}$\\
        Carcinogenesis & 22372 & 142 & 4 & 15 & $\mathcal{ALC(D)}$\\
        Mutagenesis & 14145 & 86 & 5 & 6 & $\mathcal{AL(D)}$\\
        % Pyrimidine & 74 & 1 & 0 & 27 & $\mathcal{AL(D)}$\\
        % Hepatitis & 6812 & 14 & 5 & 12 & $\mathcal{ALE(D)}$\\ 
        % Lymphography &  148 &   19 & 0 & 0 & $\mathcal{AL}$\\ 
        % Mammographic  & 975 & 33 & 3 & 2 & $\mathcal{AL(D)}$\\
        Biopax  & 323 & 28 & 19 & 30 & $\mathcal{ALCHF(D)}$\\
        % Trains  & ? & ? & ? & ? & ?\\
        \bottomrule
    \end{tabular}
\label{tab:datasets}
\end{table}

%% file: tables/single_results.tex
\begin{table*}[htb]
    \caption{Results on benchmark datasets.
    F1, Acc, T and Exp. denote the length of predicted class expression, the F1-score, the accuracy, runtime in seconds, and number of class expression tested respectively. $\dagger$ stands for no solution found and by the respective approach. $*$ indicates that respective value is not reported in DL-Learner.}
    \centering
    \small
    % RESULTS ARE double checked.
    \scalebox{0.80}{%
    \begin{tabular}{lccccccccccccccccccccccc}
    \toprule 
     \textbf{Dataset} && \multicolumn{4}{c}{\textbf{\approach}}&& \multicolumn{3}{c}{\textbf{CELOE}} && \multicolumn{4}{c}{\textbf{OCEL}}&& \multicolumn{4}{c}{\textbf{ELTL}}\\ 
    \cmidrule{3-6} \cmidrule{8-11} \cmidrule{13-16} \cmidrule{18-21} 
                            %%%%%%%%%%%%% DRILL %%%%%%%%%%%%%%%%   %%%%%%%%%%%%% CELOE %%%%%%%%%%%%%%%%%%%   %%%%%%%%%%%%%% OCEL %%%%%%%%%%%%%%%%%%%%%%%%%%%   %%%%%%%%%%%%%%%% ELTL %%%%%%%%%%%%%%%%%%%%%%%%
                             && F1     &  Acc. & T      & Exp.   && F1     & Acc.   & T         & Exp.     && F1        & Acc.          & T     & Exp.       && F1               & Acc.          & T        & Exp. \\
    \textbf{Family}          && 0.960  & 0.950 & 1.23   &  2168  && 0.970  & 0.970  & 3.63      & 646      &&  *        & .940          & 6.12  & 2756       && .960             & .950         & 3.39      & *    \\
    \textbf{Carcinogenesis}  && 0.710  & 0.560 & 4.23   &  305   && 0.710  & 0.560  & 21.14     & 230      &&$\dagger$  & $\dagger$     & 23.49 & 802        && .710             & .570          & 22.14    & *    \\
    \textbf{Mutagenesis}     && 0.700  & 0.540 & 3.04   &  3941  && 0.700  & 0.540  & 13.90     & 135      &&$\dagger$  & $\dagger$     & 13.21 & 4023       && .700             & .540          & 13.21    & *    \\
    \bottomrule
    \end{tabular}}
    \label{table:results_small}
 \end{table*}

%% file: tables/family_results.tex
\begin{table*}[!t]
    \caption{Results of single learning problems on the Family benchmark dataset. $\dagger$ stands for no solution found by the respective approach. L, F1, Acc and T denote the length of predicted class expression, the F1-score of prediction, the accuracy of prediction and runtime in seconds, respectively.}
    \centering
    \small
        \scalebox{0.90}{%

    \begin{tabular}{lccccccccccccccccccccc}
    \toprule 
      \textbf{Expression}& \multicolumn{4}{c}{\textbf{\approach}}&& \multicolumn{4}{c}{\textbf{CELOE}}&& \multicolumn{4}{c}{\textbf{OCEL}}&& \multicolumn{4}{c}{\textbf{ELTL}}\\ 
    \cmidrule{2-5} \cmidrule{7-10} \cmidrule{12-15} \cmidrule{17-20} 
                               &L      & F1    & Acc   & T     &&L      & F1    & Acc   & T     &&L       &F1    & Acc   &T            &&L    & F1    & Acc   & T \\
    \textbf{Aunt}              & 6     & 0.83  & 0.79  & 3.3   && 6     & 0.83  & 0.79  & 5.7   &&16      & *    & 1.00  & 5.8         && 1   & .800  & 0.76  & 2.8\\
    \textbf{Brother}           & 1     & 1.00  & 1.00  & 0.2   && 1     & 1.00  & 1.00  & 2.9   &&1       & *    & 1.00  & 5.8         && 5   & 1.00  & 1.00  & 3.8\\
    \textbf{Cousin}            & 4     & 0.73  & 0.65  & 2.9   && 5     & 0.79  & 0.74  & 5.9   &&21      & *    & 1.00  & 6.2         && 1   & 0.66  & 0.50  & 3.0\\
    \textbf{Daughter}          & 1     & 1.00  & 1.00  & 0.2   && 1     & 1.00  & 1.00  & 2.9   &&1       & *    & 1.00  & 5.9         && 3   & 1.00  & 1.00  & 2.9\\
    \textbf{Father}            & 1     & 1.00  & 1.00  & 0.2   && 1     & 1.00  & 1.00  & 3.0   &&1       & *    & 1.00  & 6.0         && 3   & 1.00  & 1.00  & 3.0\\
    \textbf{Granddaughter}     & 1     & 1.00  & 1.00  & 0.2   && 1     & 1.00  & 1.00  & 3.1   &&1       & *    & 1.00  & 5.3         && 1   & 1.00  & 1.00  & 2.9\\
    \textbf{Grandfather}       & 1     & 1.00  & 1.00  & 0.2   && 1     & 1.00  & 1.00  & 2.9   &&1       & *    & 1.00  & 5.7         && 1   & 1.00  & 1.00  & 3.0\\
    \textbf{Grandgranddaughter}& 1     & 1.00  & 1.00  & 0.2   && 1     & 1.00  & 1.00  & 2.9   &&1       & *    & 1.00  & 5.9         && 7   & 1.00  & 1.00  & 3.01\\
    \textbf{Grandgrandfather}  & 1     & 0.94  & 0.94  & 3.2   && 5     & 1.00  & 1.00  & 3.0   &&5       & *    & 1.00  & 5.8         && 7   & 1.00  & 1.00  & 3.7\\
    \textbf{Grandgrandmother}  & 9     & 0.94  & 0.94  & 3.3   && 5     & 1.00  & 1.00  & 3.1   &&5       & *    & 1.00  & 5.9         && 7   & 1.00  & 1.00  & 3.7\\
    \textbf{Grandgrandson}     & 1     & 0.92  & 0.92  & 3.5   && 5     & 1.00  & 1.00  & 5.7   &&5       & *    & 1.00  & 6.6         && 7   & 1.00  & 1.00  & 3.1\\
    \textbf{Grandmother}       & 1     & 1.00  & 1.00  & 0.2   && 1     & 1.00  & 1.00  & 2.8   &&1       & *    & 1.00  & 5.9         && 1   & 1.00  & 1.00  & 3.1\\
    \textbf{Grandson}          & 1     & 1.00  & 1.00  & 0.2   && 1     & 1.00  & 1.00  & 2.8   &&1       & *    & 1.00  & 6.0         && 1   & 1.00  & 1.00  & 3.0\\
    \textbf{Mother}            & 1     & 1.00  & 1.00  & 0.2   && 1     & 1.00  & 1.00  & 2.9   &&1       & *    & 1.00  & 5.9         && 5   & 1.00  & 1.00  & 3.1\\
    \textbf{PersonWithASibling}& 1     & 1.00  & 1.00  & 0.2   && 1     & 1.00  & 1.00  & 2.8   &&1       & *    & 1.00  & 7.0         && 1   & 1.00  & 1.00  & 3.1\\
    \textbf{Sister}            & 1     & 1.00  & 1.00  & 0.2   && 1     & 1.00  & 1.00  & 2.8   &&1       & *    & 1.00  & 5.8         && 5   & 1.00  & 1.00  & 3.0\\
    \textbf{Son}               & 1     & 1.00  & 1.00  & 0.2   && 1     & 1.00  & 1.00  & 3.0   &&1       & *    & 1.00  & 5.7         && 3   & 1.00  & 1.00  & 2.9\\
    \textbf{Uncle}             & 6     & 0.90  & 0.89  & 2.9   && 6     & 0.90  & 0.89  & 5.9   &&$\dagger$&$\dagger$ & $\dagger$& 5.8 && 1   & 0.88  & 0.87  & 2.9\\
    \bottomrule
    \end{tabular}}
    \label{table:family_results_detail}
 \end{table*}

%% file: tables/main_results.tex
\begin{table}[htb]
    \caption{Results on automatically generated learning problems on four benchmark datasets. \#LP stands for the number learning problems.}
    \centering
    \small
    \scalebox{0.82}{%
    \begin{tabular}{lcccccccccccccccccccc}
    \toprule 
     \textbf{Dataset} & \textbf{\#LP}&\multicolumn{4}{c}{\textbf{\approach}}&& \multicolumn{4}{c}{\textbf{CELOE}} && \multicolumn{4}{c}{\textbf{OCEL}}&& \multicolumn{4}{c}{\textbf{ELTL}}\\ 
    \cmidrule{3-6} \cmidrule{8-11} \cmidrule{13-16} \cmidrule{18-21} 
                            &      & F1    &  Acc.  & T    & Exp    && F1    & Acc.  & T      & Exp    && F1     & Acc.     & T       & Exp        && F1     & Acc.      & T      & Exp \\
    \textbf{Family}         & 74   & 1.00   & 1.00  & 1.1  & 32.2   && 1.00  & 1.00  & 3.6    & 14.2   &&  *     & 1.00     & 6.20    & 2403.0     && 1.00     & 1.00      & 3.52   & *\\
    \textbf{Carcinogenesis} & 100  & 1.00   & 1.00  & 2.2  & 47.5   && 1.00  & 1.00  & 17.3   & 16.7   &&  *     & 1.00     & 20.62   & 5876.0     && 1.00     & 1.00      & 18.6   & *\\
    \textbf{Mutagenesis}    & 100  & 1.00   & 1.00  & 1.4  & 267.8  && 1.00  & 1.00  & 10.0   & 147.9  &&  *     & 0.98     & 12.90   & 3867.4     && 0.97     & 0.97    & 10.2   & *\\
    \textbf{Biopax}         & 96   & 1.00   & 1.00  & 1.1  & 39.8   && 0.99  & 0.99  & 3.7    & 55.24  &&  *     & 1.00     & 6.74    & 5691.2     && 0.99     & 0.98     & 3.70   & *\\
    \bottomrule
    \end{tabular}}
    \label{table:results_on_generated_lps}
 \end{table}